\documentclass[letterpaper]{article} 
\usepackage{aaai2027}  
\nocopyright  

\usepackage[hyphens]{url}  
\usepackage{graphicx} 
\usepackage{natbib}  
\usepackage{caption} 
\usepackage{amsmath}
\usepackage{amssymb}
\usepackage{amsfonts}
\usepackage{algorithm}
\usepackage{algpseudocode}
\usepackage{pifont}
\newcommand{\xmark}{\ding{55}}
\usepackage{multirow}
\usepackage{booktabs}
\usepackage{caption}
\usepackage{array}

\usepackage{tabularx}
\usepackage{array}
\usepackage[table]{xcolor}

\newcolumntype{Y}{>{\centering\arraybackslash}X}

\usepackage{makecell}

\usepackage{etoolbox}

\newcommand{\TeaserFigure}{%
\par\vspace{-0.6em}
\noindent\makebox[\textwidth][c]{%
    \includegraphics[width=1\textwidth]{figures/teaser.pdf}%
}
\par\vspace{-1em}
\captionof{figure}{
EditMod performs training-free and inversion-free autoregressive image editing by updating an evolving source state with cross-conditional next-scale prediction differences.
}
\label{fig:teaser}
\par\vspace{0.6em}
}
\makeatletter
\g@addto@macro\@maketitle{\TeaserFigure}
\makeatother

\title{Model the Edit, Not the Image: Visual Autoregressive Editing from a Source-Centric Perspective}

\author{
\large\normalfont
Hongyi Fang\textsuperscript{\rm 1},
Chuwen Xie\textsuperscript{\rm 1},
Benjia Zhou\textsuperscript{\rm 1}\textsuperscript{\textdagger},
Yu-Xuan Qiu\textsuperscript{\rm 1},
Chenggong Hu\textsuperscript{\rm 2},\\
Zhibin Wang\textsuperscript{\rm 3},
Chao Chen\textsuperscript{\rm 3},
Jianbin Qin\textsuperscript{\rm 1,\rm 4},
Rui Mao\textsuperscript{\rm 4}
}
\affiliations{
\small
\textsuperscript{\rm 1}Beijing Institute of Technology, Zhuhai
\quad
\textsuperscript{\rm 2}Zhejiang University
\quad
\textsuperscript{\rm 3}Tencent
\quad
\textsuperscript{\rm 4}Shenzhen University
}

\begin{document}

\maketitle

\begingroup
\renewcommand{\thefootnote}{\fnsymbol{footnote}}
\footnotetext[2]{Corresponding author.}
\endgroup
\begin{abstract}

Next-scale visual autoregressive models (VARs) have emerged as a powerful generative paradigm, producing high-quality images through efficient coarse-to-fine prediction. However, their potential for text-guided image editing remains largely underexplored. Existing training-free VAR editing approaches often formulate editing as target-conditioned regeneration guided or constrained by the source image, and may rely on inversion, test-time optimization, attention control, or user-provided masks. This generation-centric formulation does not fully exploit the multiscale source representations provided by VARs and may introduce additional computation or intervention. We instead take a source-centric perspective on VAR editing, in which the encoded source image tokens serve as the primary visual state and the editing process focuses on condition-induced changes. Based on this perspective, we propose \textbf{EditMod}, which compares source- and target-conditioned predictions under a shared autoregressive context, treats their difference as a scale-wise editing direction, and applies it as a residual update to source tokens at selected scales. Experiments show that EditMod achieves leading source-image fidelity while maintaining strong text alignment, and completes end-to-end editing of a 1K image in only 1.57 seconds on a single A100 GPU without per-image preparation.

\end{abstract}

\begin{figure}[t]
    \centering
    \includegraphics[width=\linewidth]{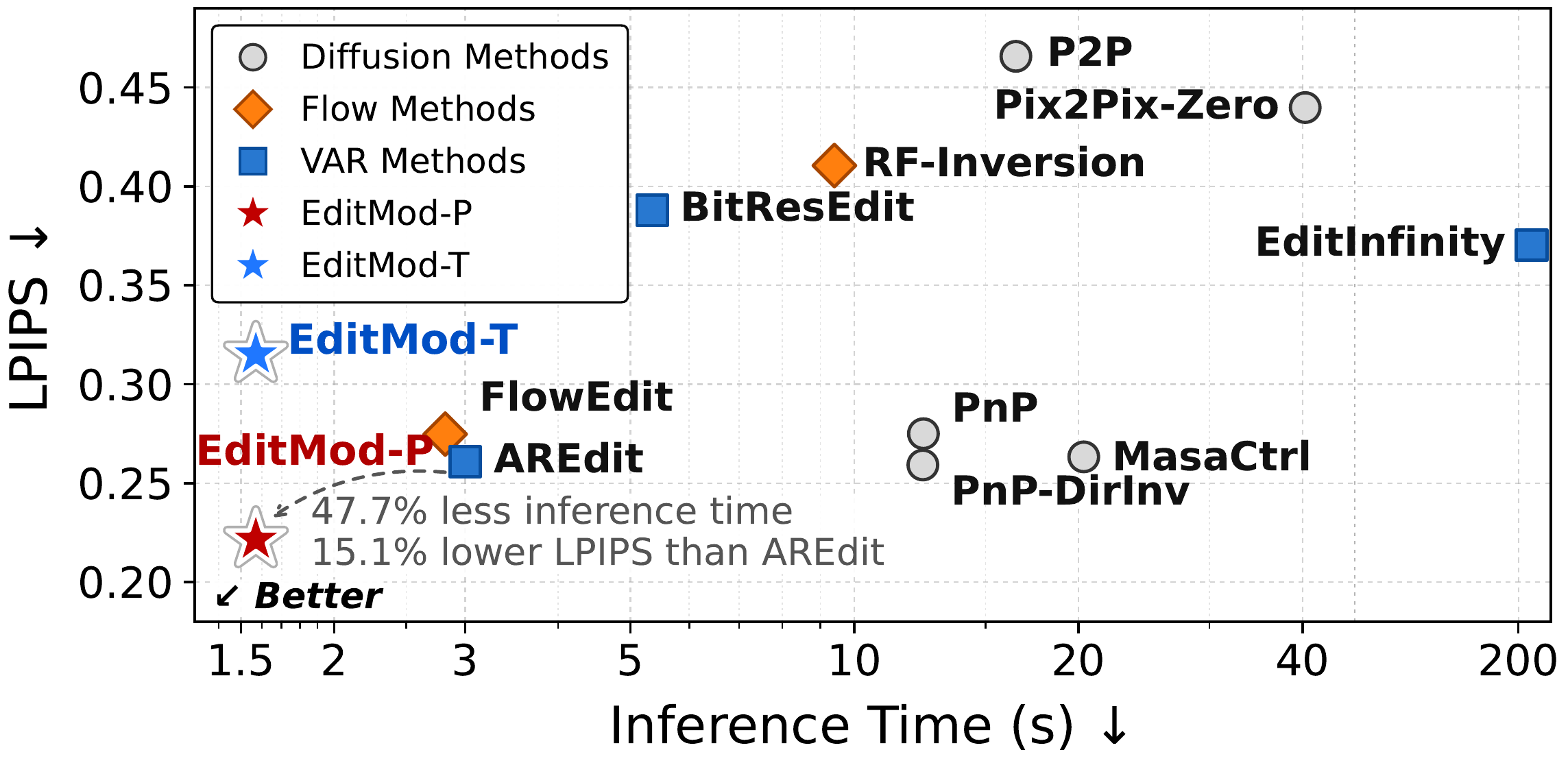}
    \caption{EditMod achieves the best inference efficiency and strong source-image fidelity among the compared methods.}
    \label{fig:intro}
    \vspace{-0.5em}
\end{figure}

%

\begin{figure}[t]
    \centering
    \includegraphics[width=0.98\linewidth]{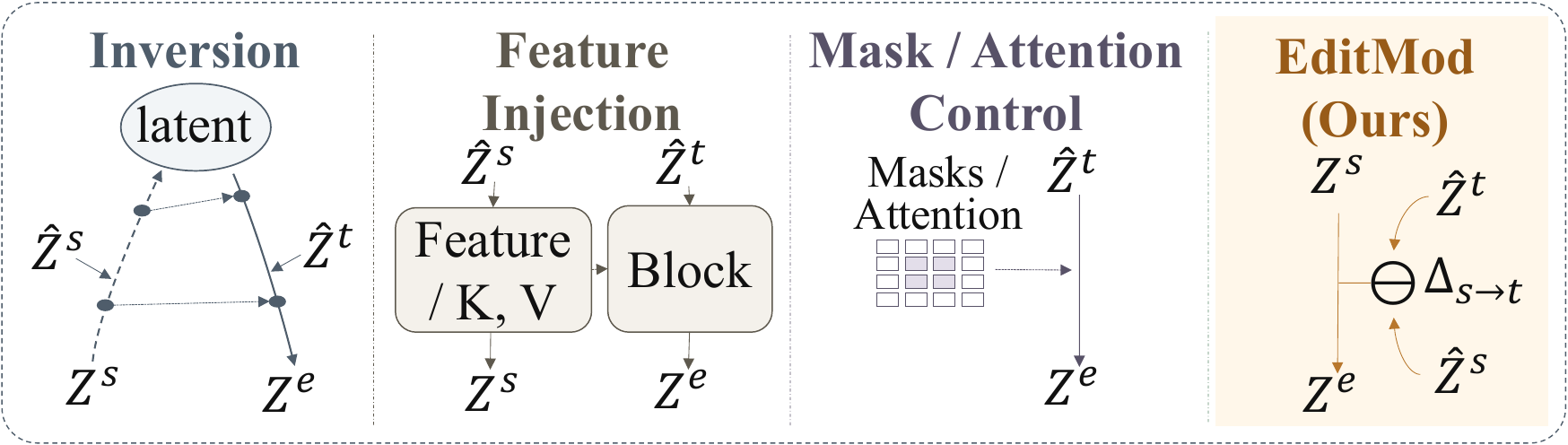}
    \caption{
    Comparison between EditMod and representative image-editing paradigms.
    }
    \label{fig:paradigm_comparison}
\end{figure}

\section{Introduction}

Next-scale visual autoregressive models (VARs) have emerged as an efficient generative paradigm that synthesizes high-quality images through discrete coarse-to-fine prediction~\cite{tian2024var,han2025infinity}. 
In contrast to diffusion models, whose sampling dynamics evolve a continuous noisy state, next-scale VARs represent an image as a hierarchy of tokens aligned with successive generation scales. 
This structural alignment provides a natural basis for \textbf{training-free} text-guided image editing: the desired semantic change may be introduced progressively across scales while preserving source information encoded at the corresponding resolutions. The key challenge is therefore not simply to regenerate an image under a new textual condition, but to modify condition-dependent content without unnecessarily altering the remaining source representation.

Despite this opportunity, most existing training-free editors remain centered on a target-conditioned generation path, treating the source image primarily as an auxiliary constraint, as shown in Figure~\ref{fig:paradigm_comparison}. Specifically, they preserve source content by recovering and following a source generation
trajectory~\cite{mokady2023null}, injecting source features~\cite{tumanyan2023pnp} or attention Key/Value states into the
target branch~\cite{cao2023masactrl}, or spatially restricting target-conditioned updates through masks~\cite{couairon2023diffedit} or attention control~\cite{hertz2023prompt}. 
At a conceptual level, this broad family constructs an edited state around a target-conditioned generation state $\widehat{Z}^{\,t}$ and then constrains or corrects it using source information $\mathcal{C}_s$, which summarizes the constraints induced by the source visual state $Z^s$ and source text condition $c_s$: 
\begin{equation}
    \widehat{Z}^{\,t} + \mathcal{C}_s \rightarrow Z^e.
\end{equation}
Consequently, content shared by the source and target conditions may still pass through the target-generation process rather than being retained directly, which can expose unchanged regions to drift and underuse the scale-aligned source representations naturally provided by VARs.

This raises a fundamental question: \emph{Rather than constraining
target-conditioned regeneration with source information, can VAR editing remain
directly anchored to the source representation and model only the predictive
change induced by the target condition?} To this end, we adopt a source-centric
formulation that retains the encoded source representation $Z^s$ as the state
being edited and constructs the edited representation as
\begin{equation}
Z^s + \Delta_{s\rightarrow t} \rightarrow Z^e.
\label{eq:source_centric}
\end{equation}
This formulation changes the basis of editing: instead of regenerating a target-conditioned state and constraining it with source information, it directly preserves the scale-aligned source representation and models only the requested semantic change $\Delta_{s\rightarrow t}$. The multi-scale source tokens of VARs provide a natural basis for this formulation, serving as scale-wise anchors while source-to-target conditional prediction differences provide the principal editing signal.

Building on this formulation, we propose \textbf{EditMod}, a training-free editing method that employs a scale-aware strategy tailored to the coarse-to-fine dynamics of VARs. At coarse scales, where tokens primarily determine global structure and spatial layout, EditMod pre-fills the source tokens to anchor the original composition. At intermediate scales, it estimates the editing direction from the difference between target- and source-conditioned predictions and applies it as a residual update to the scale-matched source representation. At fine scales, it performs target-conditioned autoregressive sampling from the edited context, allowing the model to synthesize local details consistent with the requested change. As shown in Figure~\ref{fig:intro}, EditMod achieves the best inference efficiency while maintaining strong source-image fidelity among the compared methods.

Our main contributions are summarized as follows:
\begin{itemize}
     \item We formulate VAR editing around the encoded source-image state $Z^s$
    and model the requested semantic change as a source-to-target residual
    $\Delta_{s\rightarrow t}$, rather than treating source information only as a
    constraint on a fully target-conditioned state.
    
    \item We propose \textbf{EditMod}, a mask-free, training-free VAR editor that pre-fills coarse source tokens, applies shared-context target--source prediction displacements to scale-matched source representations at intermediate scales, and samples target-consistent details at fine scales.
    
    \item Experiments demonstrate that EditMod achieves leading source-image fidelity while maintaining strong text alignment, and offers the best preparation-free editing efficiency, editing a 1K image in 1.57 seconds on a single A100 GPU.
\end{itemize}

\begin{figure*}[t]
\centering
\includegraphics[width=\textwidth]{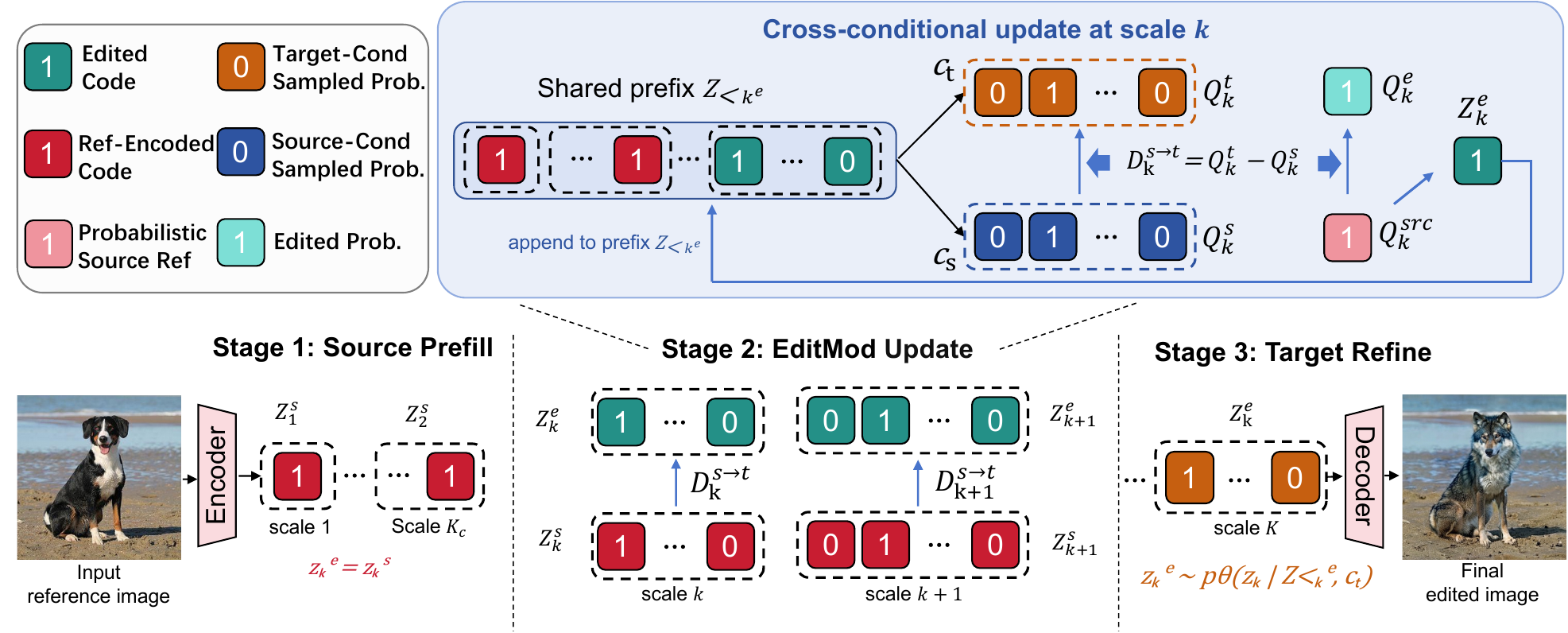}
\caption{
Overview of EditMod. At selected autoregressive scales, EditMod compares source- and target-conditioned next-scale predictions from the same edited prefix, applies the resulting displacement to the source image representation, and recursively carries the updated state forward.
}
\label{fig:method}
\vspace{-0.5em}
\end{figure*}

\section{Related Work}

\subsection{Training-Free Text-Guided Image Editing}

Training-free image editing mainly preserves source content through inversion, feature or attention reuse, and spatial constraints. Null-text Inversion optimizes the inversion trajectory for real-image reconstruction~\cite{mokady2023null}; Prompt-to-Prompt controls cross-attention to preserve spatial layouts~\cite{hertz2023prompt}; Plug-and-Play injects source features and self-attention states~\cite{tumanyan2023pnp}; MasaCtrl reuses source keys and values in the target branch~\cite{cao2023masactrl}; and DiffEdit restricts updates with an automatically derived mask~\cite{couairon2023diffedit}. Despite their different implementations, these methods generally follow a target-conditioned generation path and use source trajectories, features, attention, or masks to preserve unchanged content. FlowEdit instead reformulates inversion followed by regeneration as direct transport from the source distribution to the target distribution~\cite{kulikov2025flowedit}.
Visual autoregressive editing largely inherits these strategies. AREdit caches source tokens and predictive distributions to constrain target sampling~\cite{wang2025aredit}, while VARIN and EditInfinity recover source generation states through discrete noise inversion and per-image optimization, respectively~\cite{dao2025varin,wang2025editinfinity}. BitResEdit introduces the source prompt as an additional negative condition to strengthen target-conditioned generation and uses a spatial mask to localize the edit~\cite{zhang2026bitresedit}. Overall, existing VAR editors remain centered on target-conditioned generation constrained by source information, while direct modeling of scale-wise editing changes remains limited.

\subsection{Visual Autoregressive Generation}

Visual autoregressive generation evolved from raster-order pixel prediction~\cite{oord2016pixelcnn} to discrete latent-token modeling with VQ-VAE and VQGAN~\cite{oord2017vqvae,esser2021taming}. MaskGIT enables iterative parallel decoding through masked prediction~\cite{chang2022maskgit}, while LlamaGen demonstrates the scalability of next-token Transformers for image generation~\cite{sun2024llamagen}.
VAR reformulates visual autoregression as coarse-to-fine next-scale prediction, generating a complete token map in parallel at each scale~\cite{tian2024var}. Infinity extends this paradigm to high-resolution synthesis with multi-scale bitwise residual representations~\cite{han2025infinity}. This discrete scale hierarchy organizes visual information from global structure to local detail, providing a natural basis for retaining and modifying multi-scale source representations.

\begin{figure*}[t]
    \centering
    \includegraphics[width=0.99\textwidth]{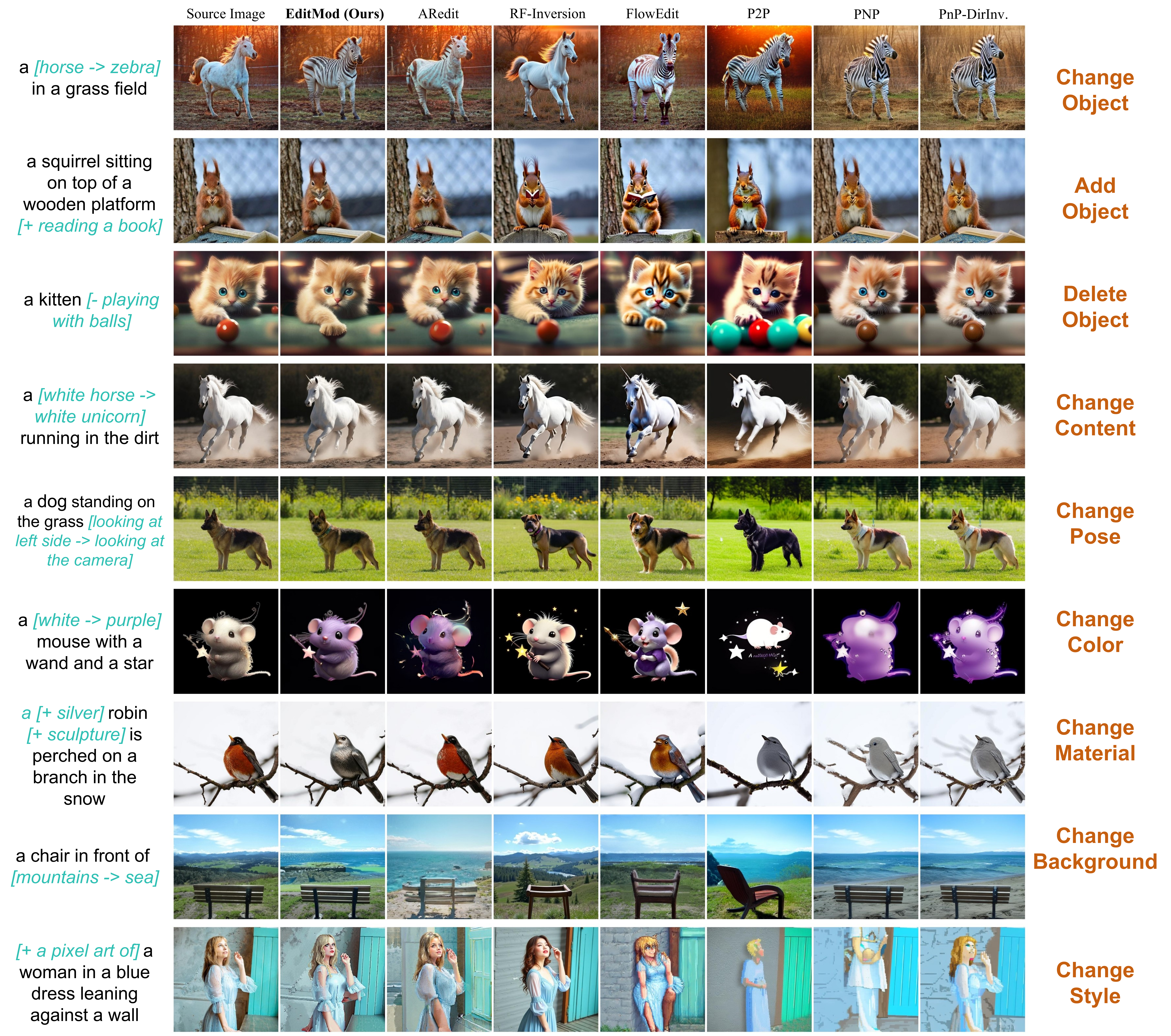}
        \vspace{-1em}
    \caption{Qualitative comparison on PIE-Bench. EditMod is consistently effective across diverse editing tasks.}
    \label{fig:qualitative}
    \vspace{-1em}
\end{figure*}

\begin{figure*}[t]
    \centering
    \includegraphics[width=\textwidth]{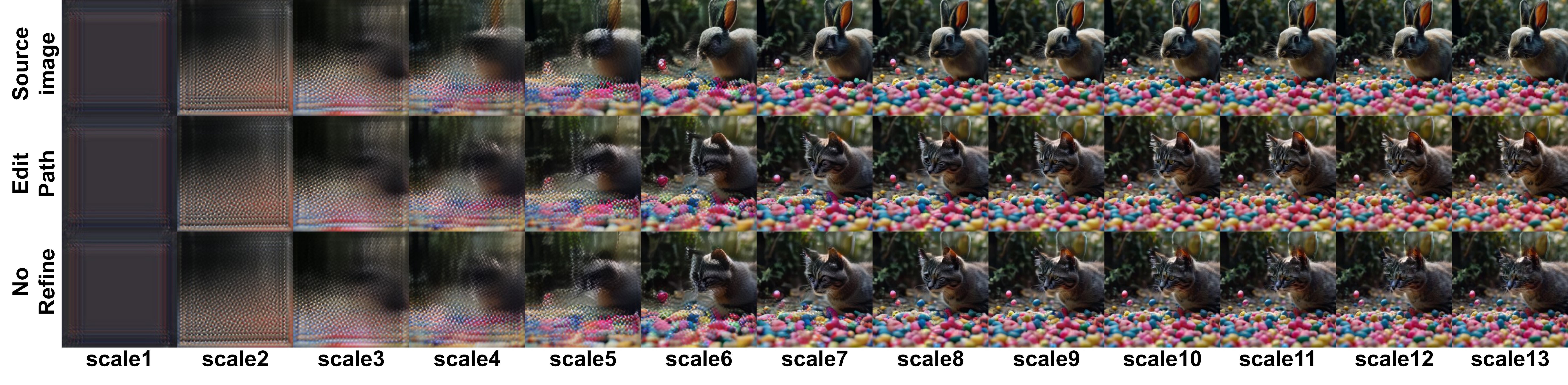}
    \caption{Scale-wise editing trajectories of the source image, EditMod, and EditMod without tail refinement.}
    \label{fig:panels}
\end{figure*}

\section{Method}
The overall editing pipeline is illustrated in Figure~\ref{fig:method}.
\subsection{Preliminaries}

Next-scale visual autoregressive modeling represents an image as a sequence of multi-scale residual maps with progressively increasing spatial resolutions. Given a source image $x^s$, the encoder produces
\begin{equation}
Z^s=\operatorname{Enc}(x^s)=\{z_1^s,\ldots,z_K^s\},
\qquad
z_k^s\in\mathbb{R}^{h_k\times w_k\times d},
\label{eq:encode}
\end{equation}
where $k$ and $K$ denote the scale index and the total number of scales, respectively, and $h_k$, $w_k$, and $d$ denote the height, width, and feature dimension at scale $k$. The model autoregressively predicts scales while generating all spatial positions within each scale in parallel:
\begin{equation}
p_\theta(z_k\mid Z_{<k},c)
=
\prod_{i=1}^{h_kw_k}
p_\theta(z_{k,i}\mid Z_{<k},c),
\label{eq:next_scale_prediction}
\end{equation}
where $Z_{<k}$ denotes the preceding scales, $c$ is the text condition, and $i$ indexes spatial positions. The image can then be reconstructed by progressively upsampling and accumulating the residual maps.
Our method is built upon Infinity, which adopts binary spherical quantization (BSQ) to enlarge the discrete vocabulary. Given a $d$-dimensional continuous feature $u$, BSQ maps it to a binary code on the unit sphere:
\begin{equation}
b
=
\frac{1}{\sqrt d}\operatorname{sign}(u)
\in
\frac{1}{\sqrt d}\{-1,+1\}^{d},
\label{eq:bsq}
\end{equation}
where $b$ denotes the quantized binary representation and $d$ is its dimensionality, yielding an effective vocabulary size of $2^d$. This representation also enables editing in the bit-wise probability space.

\subsection{Source-Centric Edit Modeling}

Following the source-centric formulation in the Introduction, we treat
the encoded source representation as the state to be edited. At scale
$k$, the edit takes the form
\begin{equation}
z_k^e
=
z_k^s+\Delta_{k,s\rightarrow t},
\label{eq:scale_edit}
\end{equation}
where $\Delta_{k,s\rightarrow t}$ denotes the scale-wise change induced
by switching the text condition from $c_s$ to $c_t$. The goal is to
estimate a target-consistent displacement while preserving source
content. The following analysis provides a conceptual formalization of
this mechanism; the introduced components serve as analytical terms
rather than explicitly separated model variables.

\paragraph{Ideal displacement.}
For analysis, we decompose the source representation and a valid target
representation into shared and condition-related components:
\begin{equation}
z_k^s=A_k+R_k^s,
\qquad
z_k^{t,*}=A_k+R_k^t,
\label{eq:representation_decomposition}
\end{equation}
where $A_k$ denotes content to be shared between the source and target,
and $R_k^s$ and $R_k^t$ denote source- and target-related semantics,
respectively. Under this decomposition, the ideal displacement is
\begin{equation}
\Delta_{k,s\rightarrow t}^{*}
=
z_k^{t,*}-z_k^s
=
R_k^t-R_k^s.
\label{eq:ideal_displacement}
\end{equation}
This formalizes an ideal edit as preserving shared content while
changing only the condition-dependent component.

\paragraph{Cross-conditional approximation.}
Since a target representation is unavailable, we obtain source- and
target-conditioned predictions under the same edited context
$Z_{<k}^e$:
\begin{equation}
\Phi_k^s=\Phi_k(Z_{<k}^e,c_s),
\qquad
\Phi_k^t=\Phi_k(Z_{<k}^e,c_t).
\label{eq:abstract_predictions}
\end{equation}
Because the two predictions share the same autoregressive context and
differ only in text condition, we conceptually separate their responses
as
\begin{equation}
\Phi_k^s=\hat A_k+\hat R_k^s,
\qquad
\Phi_k^t=\hat A_k+\hat R_k^t,
\label{eq:prediction_decomposition}
\end{equation}
where $\hat A_k$ represents context-shared responses and
$\hat R_k^s,\hat R_k^t$ represent condition-sensitive responses. Their
difference removes the shared term:
\begin{equation}
\widehat{\Delta}_{k,s\rightarrow t}
=
\Phi_k^t-\Phi_k^s
=
\hat R_k^t-\hat R_k^s.
\label{eq:abstract_displacement}
\end{equation}
EditMod applies this difference to the encoded source:
\begin{align}
z_k^e
&=
z_k^s+\widehat{\Delta}_{k,s\rightarrow t}
\nonumber\\
&=
A_k+R_k^s-\hat R_k^s+\hat R_k^t.
\label{eq:edit_construction}
\end{align}
When the source-conditioned response captures the source semantics,
i.e., $\hat R_k^s\approx R_k^s$, the construction approaches
\begin{equation}
z_k^e\approx A_k+\hat R_k^t.
\label{eq:plausible_target_solution}
\end{equation}
The target condition generally defines a set of valid representations
rather than a unique solution. Thus, $\hat R_k^t$ need only provide a
target-consistent realization under the current context, yielding a
plausible edited representation together with the retained $A_k$.

\paragraph{Preservation-error comparison.}
The same formalization clarifies the different preservation burdens of
generation- and source-centric editing. Conceptually, a
generation-centric editor can be modeled as
\begin{equation}
z_k^{e,\mathrm{gen}}
=
\tilde A_k^s+\hat R_k^t,
\label{eq:generation_centric_scale}
\end{equation}
where $\tilde A_k^s$ represents shared content recovered through
inversion, feature injection, attention control, masks, or related
constraints. Relative to the conceptual reference
$A_k+\hat R_k^t$, its preservation error is
\begin{equation}
\epsilon_k^{\mathrm{gen}}
=
\tilde A_k^s-A_k.
\label{eq:generation_error}
\end{equation}
Under the same reference, EditMod yields
\begin{equation}
\epsilon_k^{\mathrm{src}}
=
R_k^s-\hat R_k^s.
\label{eq:source_error}
\end{equation}
The distinction lies in what must be approximated for source
preservation. Generation-centric methods recover high-dimensional
shared content, including layout, identity, and local detail, whereas
EditMod inherits the encoded source state and estimates only the
source-conditioned residual. Source-centric modeling therefore shifts
the preservation burden from shared-content reconstruction to
condition-related difference estimation, reducing reliance on inversion
or explicit source constraints.

\subsection{Practical Implementation}

\subsubsection{Instantiation in Probability Space}

To instantiate the edit direction in Infinity, we convert the source BSQ codes into a probabilistic representation. For the label $b_{kij}^s\in\{-1,+1\}$ at scale $k$, spatial position $i$, and bit index $j$, we define
\begin{equation}
Q_{kij}^{\mathrm{src}}
=
\begin{cases}
p, & b_{kij}^s=+1,\\
1-p, & b_{kij}^s=-1,
\end{cases}
\label{eq:source_probability}
\end{equation}
where $p>0.5$ denotes the source-label confidence. This representation enables smoother updates than directly modifying discrete codes; alternative update spaces are discussed in the appendix.
Under the shared edited context $Z_{<k}^e$, the source- and target-conditioned bit probabilities are
\begin{equation}
Q_k^s=p_\theta(b_k\mid Z_{<k}^e,c_s),
\qquad
Q_k^t=p_\theta(b_k\mid Z_{<k}^e,c_t),
\label{eq:conditional_probabilities}
\end{equation}
where $b_k$ collects all BSQ bit variables at scale $k$, and $Q_k^s,Q_k^t$ are the corresponding bit-wise probability tensors. We define the edit direction as
\begin{equation}
D_k^{s\rightarrow t}=Q_k^t-Q_k^s,
\label{eq:probability_direction}
\end{equation}
and update the probabilistic source representation by
\begin{equation}
Q_k^e
=
\operatorname{clip}
\left(
Q_k^{\mathrm{src}}+D_k^{s\rightarrow t},
0,1
\right).
\label{eq:probability_edit}
\end{equation}
The edited BSQ code is then obtained as
\begin{equation}
z_k^e=\mathcal{B}(Q_k^e),
\label{eq:probability_to_code}
\end{equation}
where $\mathcal{B}(\cdot)$ samples each bit or selects its most probable value, yielding the scale-wise BSQ code map $z_k^e$. This code is appended to the edited context for subsequent-scale prediction.

\subsubsection{Coarse-to-Fine Editing Schedule}

VARs generate images from coarse to fine. Coarse scales mainly determine global structure, so source tokens anchor the original layout; intermediate scales introduce semantic changes; and fine scales correct autoregressive errors and refine local textures. We therefore adopt the following three-stage schedule:
\begin{equation}
z_k^e=
\begin{cases}
z_k^s,
& k\leq K_c,\\[2pt]
\mathcal{B}(Q_k^e),
& K_c<k\leq K_f,\\[2pt]
z_k\sim p_\theta(z_k\mid Z_{<k}^e,c_t),
& k>K_f,
\end{cases}
\label{eq:editing_schedule}
\end{equation}
where $K_c$ and $K_f$ denote the final scales for source prefilling and differential editing, respectively. The three stages reuse source BSQ codes, apply cross-conditional updates in probability space, and perform target-conditioned fine-scale sampling. Algorithm~\ref{alg:editmod} summarizes the complete procedure.

\begin{algorithm}[t]
\caption{Source-Centric Autoregressive Editing}
\label{alg:editmod}
\begin{algorithmic}[1]
\algrenewcommand{\algorithmiccomment}[1]{%
    \hfill\textcolor{gray}{\(\triangleright\) #1}}

\Require Source image $x^s$, source condition $c_s$, target condition $c_t$,
scale boundaries $K_c,K_f$, confidence $p$
\Ensure Edited image $x^e$

\State $Z^s=\{z_k^s\}_{k=1}^{K}\gets\operatorname{Enc}(x^s)$
\Comment{encoded source hierarchy}
\State $Z_{<1}^e\gets\varnothing$
\Comment{edited autoregressive prefix}

\For{$k=1,\ldots,K$}
    \If{$k\leq K_c$}
        \State $z_k^e\gets z_k^s$
        \Comment{source prefill}

    \ElsIf{$k\leq K_f$}
        \State $Q_k^{\mathrm{src}}
        \gets\operatorname{Prob}(z_k^s;p)$
        \State $Q_k^s\gets p_\theta(b_k\mid Z_{<k}^e,c_s)$
        \State $Q_k^t\gets p_\theta(b_k\mid Z_{<k}^e,c_t)$
        \State $Q_k^e\gets
        \operatorname{clip}
        \left(
        Q_k^{\mathrm{src}}+Q_k^t-Q_k^s,0,1
        \right)$
        \State $z_k^e\gets\mathcal{B}(Q_k^e)$
        \Comment{source-anchored differential edit}

    \Else
        \State $z_k^e\sim p_\theta(z_k\mid Z_{<k}^e,c_t)$
        \Comment{target-conditioned refinement}
    \EndIf

    \State $Z_{\leq k}^e\gets(Z_{<k}^e,z_k^e)$
\EndFor

\State $x^e\gets\operatorname{Dec}(Z_{\leq K}^e)$
\State \Return $x^e$

\end{algorithmic}
\end{algorithm}

\begin{table*}[t]
\centering

\resizebox{\textwidth}{!}{
\begin{tabular}{lc|ccc|c|ccccc}
\hline
Method & Base Model & Attn. Ctrl. & Inversion & GT Mask
& Time (s) $\downarrow$
& CLIP-T $\uparrow$ & CLIP-I $\uparrow$ & LPIPS $\downarrow$
& DINO $\uparrow$ & DreamSim $\downarrow$ \\
\hline

\multicolumn{11}{c}{\textbf{Text Alignment}} \\
\hline

FlowEdit
& Flow & \xmark & \xmark & \xmark
& \underline{2.82} & \underline{0.3232} & \underline{0.8695}
& \textbf{0.2748} & \underline{0.7505} & \underline{0.2279} \\

BitResEdit
& VAR & \xmark & \xmark & \checkmark
& 5.35 & \textbf{0.3234} & 0.8551
& 0.3879 & 0.6906 & 0.3025 \\

EditInfinity
& VAR & \xmark & \checkmark & \checkmark
& 212.31 & 0.3167 & 0.8664
& 0.3703 & 0.7188 & 0.2742 \\

\textbf{EditMod-T (Ours)}
& VAR & \xmark & \xmark & \xmark
& \textbf{1.57} & 0.3181 & \textbf{0.8762}
& \underline{0.3046} & \textbf{0.7889} & \textbf{0.1958} \\

\hline
\multicolumn{11}{c}{\textbf{Source Preservation}} \\
\hline

P2P
& Diffusion & \checkmark & \checkmark & \xmark
& 16.48 & 0.3059 & 0.8365 & 0.4657 & 0.6515 & 0.3233 \\

Pix2Pix-Zero
& Diffusion & \checkmark & \checkmark & \xmark
& 40.33 & 0.2813 & 0.7595 & 0.4398 & 0.5480 & 0.4151 \\

MasaCtrl
& Diffusion & \checkmark & \checkmark & \xmark
& 20.33 & 0.2947 & 0.8951 & 0.2633 & 0.8088 & 0.1792 \\

PnP
& Diffusion & \checkmark & \checkmark & \xmark
& 12.38 & 0.3060 & 0.8836 & 0.2750 & 0.7959 & 0.1949 \\

PnP-DirInv.
& Diffusion & \checkmark & \checkmark & \xmark
& 12.36 & 0.3064 & 0.8903 & \underline{0.2592}
& 0.8167 & 0.1770 \\

RF-Inversion
& Flow & \xmark & \checkmark & \xmark
& 9.40 & 0.3020 & 0.8737 & 0.4105 & 0.7456 & 0.2171 \\

AREdit
& VAR & \checkmark & \xmark & \xmark
& \underline{3.00} & \underline{0.3080} & \underline{0.9104}
& 0.2609 & \underline{0.8440} & \underline{0.1376} \\

\textbf{EditMod-P (Ours)}
& VAR & \xmark & \xmark & \xmark
& \textbf{1.57} & \textbf{0.3096} & \textbf{0.9120}
& \textbf{0.2212} & \textbf{0.8641} & \textbf{0.1261} \\

\hline
\end{tabular}
}
\caption{
Quantitative comparison on PIE-Bench under text-alignment- and source-preservation-oriented settings. Best and second-best results in each block are \textbf{bolded} and \underline{underlined}, respectively.
}
\label{tab:main_results}
\end{table*}










\begin{table}[t]
\centering

\small
\renewcommand{\arraystretch}{1.08}
\setlength{\tabcolsep}{1.2pt}

\begin{tabular*}{\columnwidth}{
@{\extracolsep{\fill}}
lccccc
@{}
}
\toprule
Setting
& CLIP-T $\uparrow$
& CLIP-I $\uparrow$
& LPIPS $\downarrow$
& DINO $\uparrow$
& D-Sim $\downarrow$ \\
\midrule

w/o Prefill
& \textbf{0.3170}
& 0.8537
& 0.3570
& 0.7402
& 0.2464 \\

\textbf{w/ Prefill}
& 0.3096
& \textbf{0.9120}
& \textbf{0.2212}
& \textbf{0.8641}
& \textbf{0.1261} \\

\midrule
Rel. $\Delta$
& $-2.3\%$
& $+6.8\%$
& $-38.0\%$
& $+16.7\%$
& $-48.8\%$ \\

\bottomrule
\end{tabular*}
\caption{
Effect of coarse-scale source prefilling.
}
\label{tab:prefill_ablation}
\end{table}

\begin{table}[t]
\centering

\small
\renewcommand{\arraystretch}{1.08}
\setlength{\tabcolsep}{1.2pt}

\begin{tabular*}{\columnwidth}{
@{\extracolsep{\fill}}
lccccc
@{}
}
\toprule
Middle Operation
& CLIP-T $\uparrow$
& CLIP-I $\uparrow$
& LPIPS $\downarrow$
& DINO $\uparrow$
& D-Sim $\downarrow$ \\
\midrule

Src. Sampling
& 0.3010
& \underline{0.8678}
& \underline{0.5137}
& \underline{0.7100}
& \underline{0.2661} \\

Tgt. Sampling
& \textbf{0.3236}
& 0.8351
& 0.5237
& 0.6602
& 0.3117 \\

\midrule
\textbf{Diff. Update}
& \underline{0.3096}
& \textbf{0.9120}
& \textbf{0.2212}
& \textbf{0.8641}
& \textbf{0.1261} \\

\bottomrule
\end{tabular*}
\caption{
Comparison of intermediate-scale operations.
}
\label{tab:middle_operation}
\end{table}

\begin{figure}[t]
    \centering
    \includegraphics[width=\linewidth]{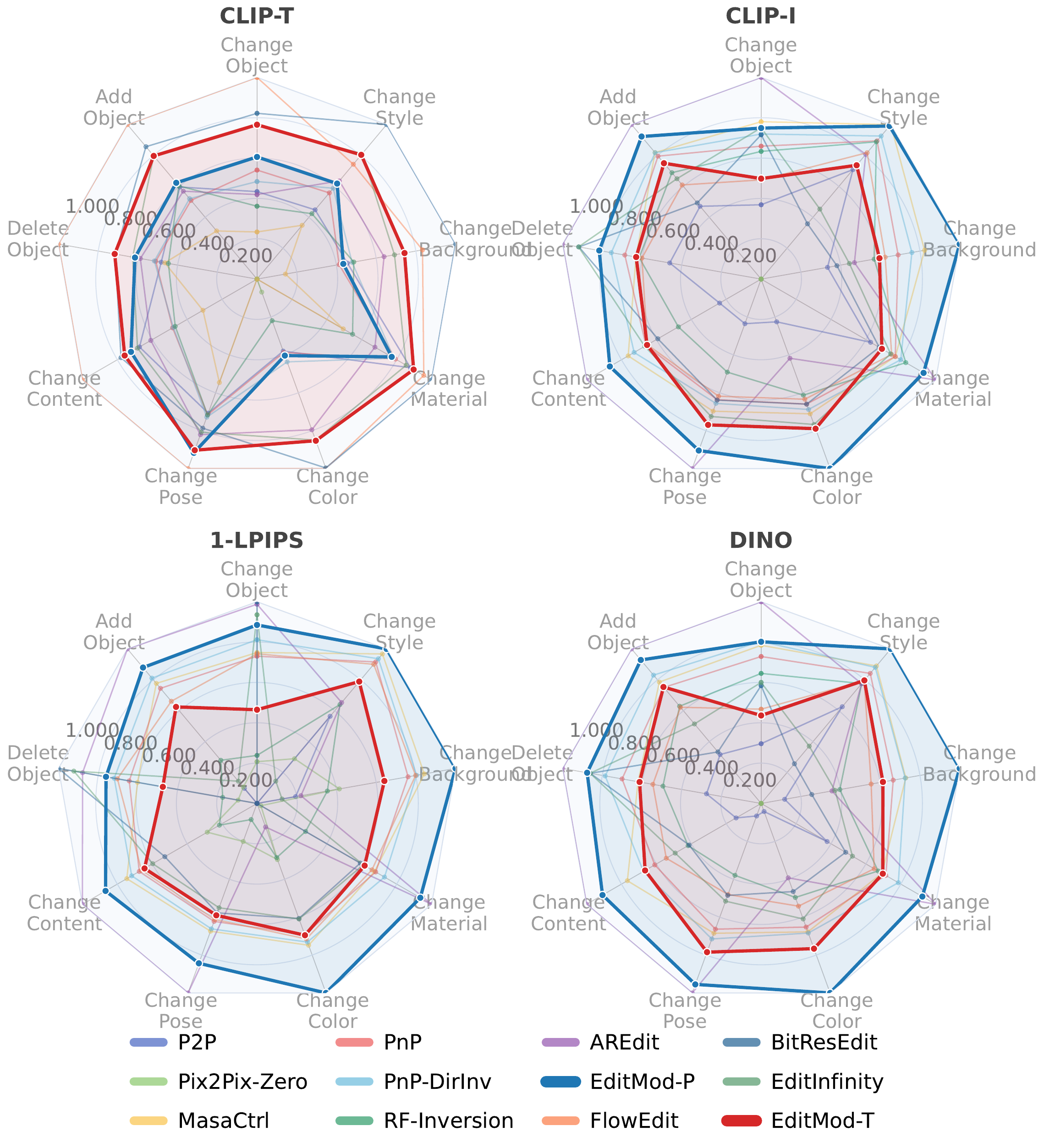}
    \caption{Task-wise normalized performance across PIE-Bench editing categories.}
    \label{fig:taskwise}
\end{figure}

\section{Experiments}

\subsection{Experimental Setup}

\paragraph{Benchmark.}
We conduct the main comparison on PIE-Bench~\cite{Xuan2023direct_inversion}, which contains 700 real source images, each paired with a source prompt and a target prompt. The benchmark covers diverse editing tasks, including object replacement, attribute modification, pose change, style transfer, background editing, and local semantic editing, allowing us to evaluate both text alignment and source preservation.

\paragraph{Implementation details.}
We use Infinity-2B~\cite{han2025infinity} as the backbone, which represents each image over
$K=13$ autoregressive scales. Here, $w$ denotes the classifier-free
guidance scale~\cite{Ho2022classifier_free_guidance}, $p$ the source-reference probability, and $K_c$ and
$K_f$ the coarse and fine scale boundaries, respectively. To evaluate
the trade-off between text alignment and source preservation, we use two
configurations: EditMod-P sets $w=8$, $p=0.99$, $K_c=3$, and $K_f=8$,
while EditMod-T sets $w=8$, $p=0.99$, $K_c=2$, and $K_f=7$.

\paragraph{Baselines.}
We compare representative training-free editors across diffusion, flow, and visual autoregressive paradigms using their standard backbones. P2P~\cite{hertz2023prompt}, Pix2Pix-Zero~\cite{Parmar2023pix2pix_zero}, MasaCtrl~\cite{cao2023masactrl}, PnP~\cite{tumanyan2023pnp}, and PnP-DirInv.~\cite{tumanyan2023pnp,Xuan2023direct_inversion} use Stable Diffusion 1.5~\cite{Rombach2021latent_diffusion}; RF-Inversion~\cite{Wang2025rf_inversion} and FlowEdit~\cite{kulikov2025flowedit} use Stable Diffusion 3~\cite{Esser2024stable_diffusion_3}; and EditMod, AREdit~\cite{wang2025aredit}, BitResEdit~\cite{zhang2026bitresedit}, and EditInfinity~\cite{wang2025editinfinity} use Infinity-2B~\cite{han2025infinity}. These backbone choices follow the default configurations of each paradigm. All methods are evaluated on the same source images and target prompts.

\paragraph{Metrics.}
We evaluate text alignment with CLIP-T~\cite{Radford2021clip} and source preservation with CLIP-I~\cite{Radford2021clip}, DINO~\cite{Caron2021dino}, LPIPS~\cite{Zhang2018lpips}, and DreamSim~\cite{Fu2023dreamsim}. Higher CLIP-T, CLIP-I, and DINO are better, while lower LPIPS and DreamSim are preferred. Efficiency is measured by end-to-end runtime per image on a single NVIDIA A100.

\subsection{Main Results}

\paragraph{Quantitative comparison.}
Existing baselines naturally occupy different regions of the
alignment--preservation trade-off: some favor text alignment, while
others prioritize source fidelity. Table~\ref{tab:main_results}
therefore reports two EditMod operating points against the corresponding
groups, with all baselines evaluated under their reported or official
settings. EditMod-T remains competitive among alignment-oriented
methods while providing stronger source preservation, whereas EditMod-P
outperforms preservation-oriented methods across all metrics. Notably,
some baselines additionally rely on user-provided or ground-truth masks,
which explicitly localize the edit and provide a strong spatial prior;
EditMod achieves these results without such supervision, training,
inversion, or attention control.

\paragraph{Qualitative comparison.}
Figure~\ref{fig:qualitative} presents representative results across diverse editing categories. EditMod consistently performs the intended edits while better preserving object identity, spatial layout, and unedited content.

\paragraph{Task-wise comparison.}
Figure~\ref{fig:taskwise} reports per-task min--max normalized results. EditMod maintains strong source preservation across most categories while remaining competitive in text alignment, demonstrating the generality of the proposed source-centric formulation.

\subsection{Ablation Studies}

\paragraph{Source prefilling.}
Table~\ref{tab:prefill_ablation} validates the role of coarse-scale
source prefilling. Reusing source tokens improves CLIP-I and DINO by
$6.8\%$ and $16.7\%$, while reducing LPIPS and DreamSim by $38.0\%$
and $48.8\%$, respectively, with only a $2.3\%$ drop in CLIP-T.
This asymmetric trade-off supports our scale-wise design: early VAR
scales mainly encode global geometry and spatial layout, so preserving
them provides a stable structural anchor while leaving sufficient
capacity for semantic changes at later scales.

\paragraph{Intermediate-scale operation.}
Table~\ref{tab:middle_operation} compares three intermediate-scale
operations. Target-conditioned sampling achieves the highest CLIP-T
but substantially harms source preservation, while source-conditioned
sampling remains weaker overall than the proposed differential update.
The differential update reaches a CLIP-I of $0.9120$ and reduces LPIPS
from above $0.51$ for both sampling variants to $0.2212$, while
retaining competitive CLIP-T. These results show that its advantage is
not a simple interpolation between source and target sampling. Instead,
comparing the two conditions under the same edited context suppresses
shared responses and transfers primarily the condition-induced change,
thereby preserving source-aligned content more effectively.

\paragraph{Tail refinement.}
Figure~\ref{fig:panels} shows that semantic changes are mainly formed at
intermediate scales, while later scales primarily refine high-frequency
details. Without tail refinement, local textures and object boundaries
remain less stable, indicating that differential updates alone are less
suited to fine-scale synthesis. Switching to a clean target-conditioned
branch at the final scales improves these details and reduces local
artifacts, while largely preserving the semantic edit already established
at the intermediate scales.

\paragraph{Additional results.}
The appendix includes sensitivity analyses of scale boundaries, CFG strength, and source-reference probability, comparisons across update spaces, and additional qualitative results and failure cases.

\section{Conclusion}
EditMod edits the encoded source state across autoregressive scales, preserving coarse structure, introducing target-consistent semantics at intermediate scales, and refining fine-scale details. Without training, inversion, masks, attention control, or per-image optimization, it shows that visual autoregressive representations provide a natural and efficient state space for source-centric editing, retaining the source state while modeling only the requested semantic change.

\section{Limitations}
Our results demonstrate the promise of source-centric modeling for visual autoregressive editing. Nevertheless, EditMod is only an initial instantiation of this paradigm, leaving alternative edit-direction and source-update formulations underexplored. Moreover, its simple and efficient three-stage schedule relies on coarse predefined boundaries, motivating more adaptive, fine-grained, and semantics-aware cross-scale control.

\bibliography{aaai2027}

\end{document}